\pdfoutput=1
\documentclass[a4paper,margin=2.5cm,heightrounded=true]{article}
\usepackage{acl}
\usepackage{times}
\usepackage{latexsym}
\usepackage[T1]{fontenc}
\usepackage[utf8]{inputenc}
\usepackage{microtype}
\usepackage{inconsolata}
\usepackage{graphicx}
\usepackage{amsmath}
\usepackage{amssymb}
\usepackage{underscore}
\usepackage[utf8]{inputenc}
\usepackage[T1]{fontenc}
\usepackage{booktabs}
\usepackage{geometry} 

\usepackage{hyperref}  
\usepackage{microtype}
\usepackage{parskip}
\usepackage{booktabs}
\usepackage{siunitx}
\usepackage{colortbl}
\usepackage{xcolor}
\newcounter{fncounter}
\newcommand{\adaptcomm}{$\pi_{\text{comm}}$}
\usepackage{babel}

\title{OSC: Cognitive Orchestration through Dynamic Knowledge Alignment\\ in Multi-Agent LLM Collaboration}

\author{
  \textbf{Jusheng Zhang$^{1}$, Yijia Fan$^{1}$, Kaitong Cai$^{1}$, Xiaofei Sun$^{2}$, Keze Wang$^{1,\dagger}$}\\
  \vspace{2em}
  $^{1}$Sun Yat-sen University \quad $^{2}$Alibaba Group\\
  \texttt{} 
  $^\dagger$Corresponding author: kezewang@gmail.com
}

\begin{document}
\maketitle
\begin{abstract}
This paper introduces OSC (Orchestrating Cognitive Synergy), a knowledge-aware adaptive collaboration framework designed to enhance cognitive synergy in multi-agent systems with large language models. While prior work has advanced agent selection and result aggregation, efficient linguistic interactions for deep collaboration among expert agents remain a critical bottleneck. OSC addresses this gap as a pivotal intermediate layer between selection and aggregation, introducing Collaborator Knowledge Models (CKM) to enable each agent to dynamically perceive its collaborators' cognitive states. Through real-time cognitive gap analysis, agents adaptively adjust communication behaviors, including content focus, detail level, and expression style, using learned strategies. Experiments on complex reasoning and problem-solving benchmarks demonstrate that OSC significantly improves task performance and communication efficiency, transforming ``parallel-working individuals'' into a ``deeply collaborative cognitive team.'' This framework not only optimizes multi-agent collaboration but also offers new insights into LLM agent interaction behaviors.
\end{abstract}
\section{Introduction}
\label{sec:introduction}

In recent years, large language models (LLMs)\cite{llama,gpt,gpt2,gpt4} have shown exceptional capabilities in tackling complex tasks, greatly advancing artificial intelligence. However, scaling a single LLM often leads to high computational costs and performance bottlenecks. Multi-agent systems (MAS)\cite{mas1,mas2,router,mas4} offer a scalable alternative by leveraging diverse agents' expertise to solve problems beyond the reach of individual models, improving cost-efficiency and unlocking LLMs' full potential. Recent research\cite{router,router1,kabb} has focused on efficient MAS collaboration, with ``dynamic expert selection'' and knowledge-aware routing frameworks effectively matching tasks to expert subsets, boosting adaptability and resource efficiency.

Moreover, ``aggregation strategies'' aim to combine multi-agent outputs into high-quality final solutions. Yet, a critical challenge remains: even with an optimal expert combination, enabling these experts to dynamically adapt their linguistic interactions—fostering shared understanding, resolving discrepancies, and producing coherent, high-quality outputs—remains a key bottleneck in MAS-LLM research.

To tackle this, we propose OSC (Orchestrating Cognitive Synergy), an end-to-end, knowledge-aware adaptive collaboration framework. OSC serves as an intermediate layer, enhancing linguistic interactions among selected experts without replacing expert selection or aggregation. In its ``inter-expert collaborative communication'' phase, each agent $e_i$ uses a dynamically learned Collaborator Knowledge Model ($CKM_i(e_j \mid Q, H_t)$) to track collaborators' cognitive states (knowledge, reasoning, task understanding via $H_t$). CKM parameters ($\theta_{\text{CKM}}, \theta_{\text{update}}$), initially pre-trained, are fine-tuned end-to-end within OSC's RL loop, tailoring them for effective collaboration. A learnable cognitive gap analysis module ($\mathcal{G}_{i,j}$) informs a policy $\pi_{\text{comm}}$, which dynamically shapes communication behavior $M_{i \to j}$ (content, style, objectives; $\Phi_i^{(t)}$ as $e_i$'s state). This enables precise information sharing, plan coordination, and conflict resolution. OSC's components adapt through task feedback, ensuring synergistic, adaptive collaboration.

OSC turns experts from "parallel workers" into a "collaborative cognitive team" through adaptive language interactions, enabling robust consensus, efficient discrepancy resolution, and optimized solutions.

The primary contributions of this work are:
\begin{itemize}
    \item OSC Framework: A knowledge-aware, end-to-end framework that enhances MAS-LLM collaboration through adaptive inter-agent linguistic interactions.
    \item Collaboration Mechanisms: Trainable components—Collaborator Knowledge Modeling (CKM), cognitive gap analysis ($\mathcal{G}_{i,j}$), and communication policies ($\pi_{\text{comm}}$)—enable dynamic information exchange and conflict resolution.
     \\
    \item Validation and Insights: OSC outperforms baselines on complex reasoning benchmarks (MATH\cite{MATH}), offering new insights into LLM-agent collaboration.
\end{itemize}
\section{Related Work}
\subsection{LLM-Driven Multi-Agent Systems}
Recent work\cite{mas11,mas22,GAM,GAM2,KAF} on LLM-based multi-agent systems (MAS) explores their potential for complex tasks by combining diverse model strengths, improving efficiency over single models. Some systems\cite{semas,semas1} simulate software development teams, assigning roles like product manager or programmer to LLM agents for collaborative task completion. Others\cite{wkmas1,wkmas2} introduce structured workflows to align with engineering practices or enable flexible agent interactions that adapt to task needs.
These approaches show promise but rely on fixed roles and protocols, lacking awareness of agents’ knowledge states or adaptive adjustments. They prioritize final task outcomes over optimizing collaboration, which our OSC framework targets.
\subsection{Agent Selection and Result Aggregation}
Agent selection and result aggregation are critical for MAS efficiency\cite{efmas1,efmas2}. Knowledge-aware routing\cite{COKE} matches tasks to agents based on capabilities, while dynamic routing\cite{chen2024frugalgpt} adjusts allocations using historical performance. Continual learning helps agents acquire new skills for better task distribution.
Aggregation methods include voting-based techniques\cite{masft}, self-assessment for response reliability\cite{uncerdebate}, and hierarchical fusion\cite{layer} for integrating varied information. These treat collaboration as a black box, neglecting interaction optimization, unlike OSC’s focus on enhancing mid-process collaboration.
\subsection{Inter-Agent Communication}
Communication enables deep collaboration. Some approaches extend chain-of-thought prompting to share reasoning, use debate frameworks\cite{debate1,debate2} to refine solutions, or standardize dialogue formats. These remain static, lacking dynamic adaptation.
Negotiation mechanisms resolve disagreements, and consensus-building techniques align diverse viewpoints, but they lack systematic knowledge modeling. Information-sharing methods, like shared memory\cite{memory} or incremental learning\cite{jovanovic2024incrementallearninglargelanguage,graziuso-etal-2024-task}, focus on transmission without considering recipients’ cognitive states.
In contrast, OSC employs Collaborator Knowledge Models (CKM) for precise cognitive state modeling, adaptive communication strategies based on cognitive gap analysis, and reinforcement learning\cite{ppo} to optimize interactions and enhance MAS collaboration.

\section{Method}
\label{sec:method}
\begin{figure*}[htbp]
    \centering
    \includegraphics[width=0.95\textwidth]{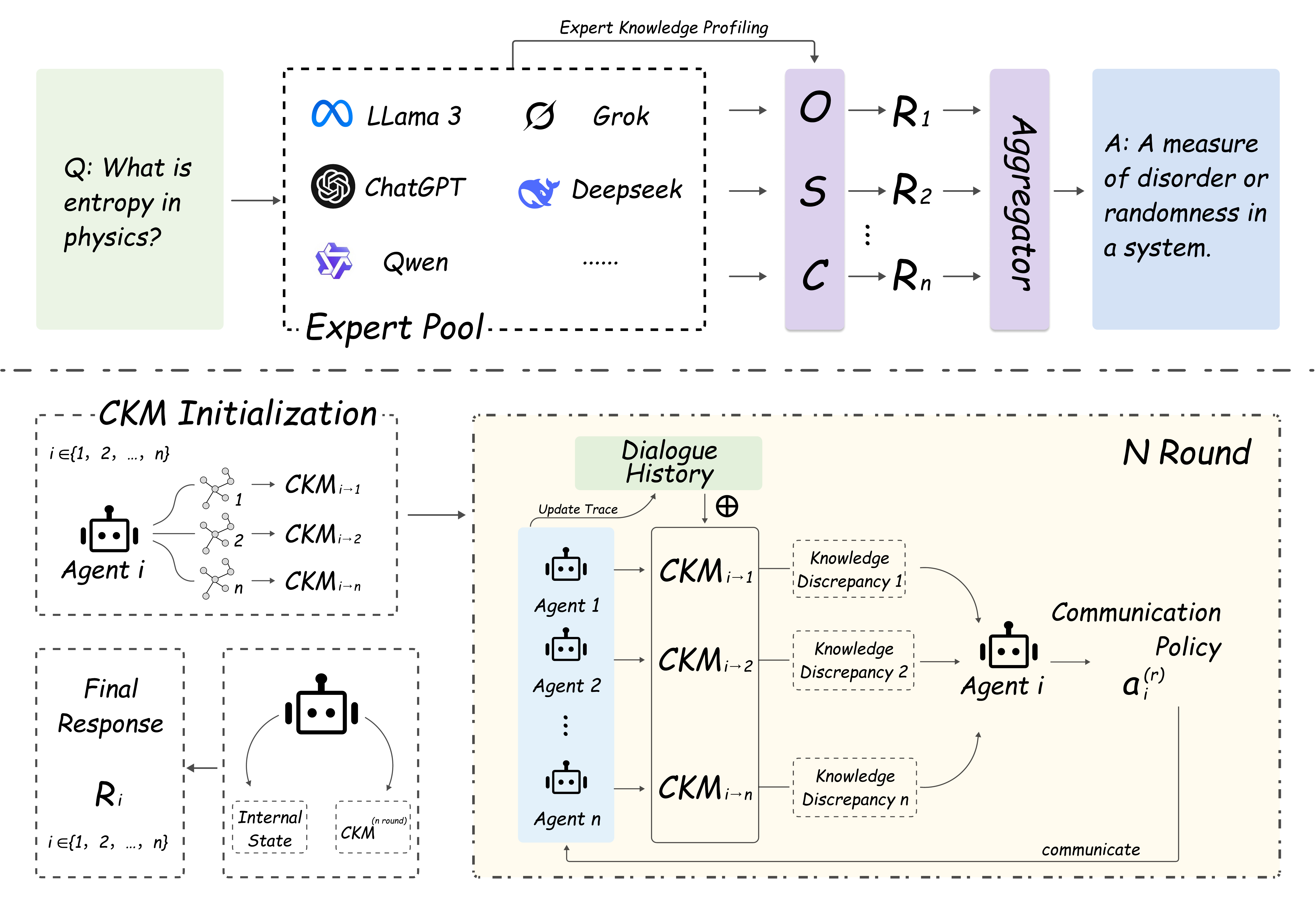}
    \caption{The OSC framework uses Collaborator Knowledge Models (CKMs)}
    \label{fig:xiaorong}
\end{figure*}
To address inefficiencies in collaborative communication within multi-agent systems (MAS) using large language models (LLMs) post-expert selection, we propose OSC (Orchestrating Cognitive Synergy). OSC introduces a structured linguistic interaction phase, transforming selected expert agents from parallel workers into a cohesive, intelligent team. This phase features dynamically learned models of agents' cognitive states and adaptive communication policies, fine-tuned end-to-end. These enable agents to perceive, reason, and respond to evolving team knowledge and intentions. Through integrated learning, agents refine solutions, resolve conflicts, and reach robust consensus before final answer aggregation, guided by a reinforcement-learned communication policy, \adaptcomm{} (see Section~\ref{ssec:adaptive_comm_strategy}).

\subsection{OSC Framework}
\label{ssec:osc_framework}
The OSC framework acts as an adaptive collaborative reasoning layer between expert selection and answer aggregation. For a query $Q$ and expert subset $\mathcal{S}_t = {e_1, \dots, e_k}$, OSC’s intelligence emerges via core, interconnected stages.
\textbf{Dynamic Collaborator Knowledge Model (CKM) and Adaptation.}
For each expert $e_i \in \mathcal{S}t$, a Collaborator Knowledge Model $CKM_i(e_j | Q, H_t)$ is created for every other expert $e_j$ ($j \neq i$). This dynamic model captures $e_i$’s evolving understanding of $e_j$’s knowledge, reasoning, confidence, and query $Q$ comprehension as dialogue $H_t$ progresses. Initialized from pre-training on large-scale dialogue corpora (Section~\ref{ssec:ckm}, Appendix~\ref{app:ckm_details_revised}), CKM parameters $\theta{\text{CKM}}$ and $\theta_{\text{update}}$ are fine-tuned end-to-end in OSC’s reinforcement learning loop. 
\textbf{Iterative Adaptive Communication.} The system engages in $N_{\text{round}}$ communication rounds (typically $N_{\text{round}}=3 \text{ to } 5$ in our experiments, a hyperparameter tuned on a development set). In each round $r \in [1, N_{\text{round}}]$:
\begin{itemize}
    \item Each expert $e_i$ (following a round-robin speaking order, though other scheduling policies can be integrated) leverages its continuously updated $CKM_{i}^{(r-1)}(e_j | Q, H^{(r-1)})$ for all collaborators $e_j$. This model is used to perform a \textbf{learned cognitive gap analysis}, yielding $\mathcal{G}_{i,j}^{(r)}$. This gap, detailed in Section~\ref{ssec:cognitive_gap}, quantifies the communicatively significant divergence between $e_i$'s internal cognitive state $\Phi_i^{(r-1)}$ (e.g., its own solution plan or understanding related to $Q$) and its CKM-derived assessment of $e_j$'s corresponding state. The function $f_{\text{gap}}$ that computes this is itself a learnable component, enabling OSC to identify discrepancies most relevant for guiding communication.
    \item Based on the matrix of identified cognitive gaps $\{\mathcal{G}_{i,j}^{(r)}\}_{j \neq i}$ across the team, expert $e_i$ employs its \textbf{adaptive communication strategy} \adaptcomm{}. This policy, optimized via reinforcement learning (PPO; details in Appendix~\ref{app:rl_details}), selects a structured, abstract communication action $a_i^{(r)}\sim\pi_{\text{comm}}(\cdot\mid
  \Phi_i^{(r-1)},\allowbreak
  \{CKM_i^{(r-1)}(e_j)\}_{j\neq i},\allowbreak
  \{\mathcal{G}_{i,j}^{(r)}\}_{j\neq i},\allowbreak
  Q,\,H^{(r-1)})$. The policy learns to map the rich, CKM-informed state to multi-faceted actions that are predicted to effectively bridge cognitive gaps and advance collective problem-solving.
    \item The abstract action $a_i^{(r)}$ encapsulates the learned communicative intent: specifically, \textit{what cognitive aspects} to address, with \textit{which collaborator(s)}, using \textit{what communication objective} (e.g., clarification, proposal, critique), and employing \textit{what interactional style} (e.g., level of detail, confidence expression). This structured directive $a_i^{(r)}$ is then verbalized into a natural language message $m_i^{(r)}$ by a generative language model, $f_{\text{LLM}}$. Importantly, $f_{\text{LLM}}$ acts as a \textbf{linguistic realization engine} conditioned on the precise, strategically determined output from OSC's learned components. OSC dictates the communicative strategy, while $f_{\text{LLM}}$ renders it into language (Section~\ref{ssec:adaptive_comm_strategy_details}, with prompt details in Appendix~\ref{app:prompt_details_revised}).
    \item All experts $e_j \in \mathcal{S}_t$ update their dialogue history $H^{(r)} = H^{(r-1)} \cup \{m_i^{(r)}\}_{i \in \mathcal{S}_t}$ and, crucially, update their respective Collaborator Knowledge Models $CKM_j^{(r)}(e_l | Q, H^{(r)})$ using the learned update mechanism $f_{\text{update}}$ (Section~\ref{ssec:ckm}). 
\end{itemize}

\textbf{Optimized Independent Contribution Generation.} Following $N_{\text{round}}$ rounds of OSC-driven communication, each expert $e_i$ generates its refined individual response $R_i$ to query $Q$. This response is conditioned on its final internal state $\Phi_i^{(N_{\text{round}})}$, which has been significantly shaped and informed by the preceding collaborative dialogue, and its comprehensive understanding of collaborators' likely final states as encoded in $CKM_i^{(N_{\text{round}})}$.

\textbf{Answer Aggregation and Propagated Collaborative Reward.} An aggregator module then combines the individual, refined contributions $\{R_i\}_{i=1}^k$ (e.g., using a learned meta-LLM aggregator or task-specific heuristics) to produce the final system output $R_{\text{final}}$. The quality of $R_{\text{final}}$ (e.g., task success, score on a benchmark) provides the primary reward signal $R_{\text{task}}$ for optimizing \adaptcomm{}. This global reward signal is also used to provide supervisory signals for the end-to-end fine-tuning of the CKM parameters ($\theta_{\text{CKM}}$, $\theta_{\text{update}}$) and the cognitive gap analysis module ($\theta_{\text{gap}}$),

\subsection{Dynamic Collaborator Knowledge Model (CKM)}
\label{ssec:ckm}
The CKM is the epistemic foundation of OSC, enabling each agent $e_i$ to construct and maintain a dynamic, internal model $CKM_i(e_j | Q, H_t)$ of each collaborator $e_j$’s evolving cognitive state relevant to the task $Q$ and the dialogue history $H_t$.

While a comprehensive ontology of cognitive features can be vast, OSC starts from a broad set of \textit{candidate cognitive dimensions} $\mathcal{C}_Q^* = \{c^*_1, c^*_2, \dots, c^*_p\}$. These can include general linguistic markers, common reasoning patterns, or task-agnostic conversational acts (examples in Appendix~\ref{app:ckm_details_revised} under "Candidate Cognitive Dimensions"). Critically, OSC does not rely on a fixed, manually selected subset of these for each task. Instead, the CKM function $f_{\text{CKM}}$ learns to \textbf{attend to and represent the most task-relevant facets} indicated by these candidate dimensions, effectively deriving a dynamic, latent cognitive state representation $\mathbf{z}_{ij}^{(t)} \in \mathbb{R}^{d_{\text{ckm}}}$ ($d_{\text{ckm}}=128$ in our setup) that is optimally conditioned on $e_j$'s behavior, the query $Q$, and history $H_t$:
\begin{equation}
\mathbf{z}_{ij}^{(t)} = f_{\text{CKM}}(e_j, Q, H_t; \theta_{\text{CKM}}) \quad 
\label{eq:ckm_state}
\end{equation}
HereArchitecture in Appendix~\ref{app:ckm_details_revised}, $\theta_{\text{CKM}}$ are the parameters of $f_{\text{CKM}}$ (typically a Transformer encoder architecture; see Appendix~\ref{app:ckm_details_revised} for model details). The learned latent vector $\mathbf{z}_{ij}^{(t)}$ implicitly encodes aspects crucial for collaboration, such as $e_j$’s evolving understanding of specific sub-problems, its confidence on particular deductions, or its awareness of specific constraints, without these needing to be explicitly predefined as rigidly structured slots. $f_{\text{CKM}}$ processes $e_j$'s utterances and interaction patterns to infer these latent attributes.
The CKM parameters $\theta_{\text{CKM}}$ and the parameters $\theta_{\text{update}}$ of the state transition function $f_{\text{update}}$ (implemented as a GRU; $d_{\text{gru}}=128$; details in Appendix~\ref{app:ckm_details_revised}):
\begin{equation}
\mathbf{z}_{ij}^{(t+1)} = f_{\text{update}}(\mathbf{z}_{ij}^{(t)}, m_j^{(t)}, Q, H_t; \theta_{\text{update}}) 
\label{eq:ckm_update}
\end{equation}
The models are initialized via pre-training on large dialogue corpora using self-supervised objectives (see Appendix~\ref{app:ckm_details_revised}). Crucially, after initialization, $\theta_{\text{CKM}}$ and $\theta_{\text{update}}$ are \textbf{continuously fine-tuned during the main reinforcement learning phase of \adaptcomm{}}. Gradients from the overall task reward $\mathcal{R}$, along with optional auxiliary losses for intermediate collaborative success (e.g., conflict resolution, plan alignment), are backpropagated to these modules. This end-to-end training enables CKM to represent collaborator states in ways that directly benefit the agent’s communication policy and task performance.
\subsection{Learned Cognitive Gap Analysis and Adaptive Communication Objectives}
\label{ssec:cognitive_gap}
Effective communication hinges on identifying and addressing the cognitive gap $\mathcal{G}_{i,j}^{(t)}$ between an expert $e_i$’s internal cognitive state $\Phi_i^{(t)}$ (e.g., its current plan embedding or understanding of $Q$) and its CKM-derived model of $e_j$’s state $\mathbf{z}_{ij}^{(t)}$. The mapping of $\Phi_i^{(t)}$ and $\mathbf{z}_{ij}^{(t)}$ into a common, comparable representational space is facilitated by learnable projection layers, which are co-trained with the CKM and \adaptcomm{} to ensure semantic alignment.

The cognitive gap function, $f_{\text{gap}}$, is itself a \textbf{learnable neural component parameterized by $\theta_{\text{gap}}$}:
\begin{equation}
\mathcal{G}_{i,j}^{(t)} = f_{\text{gap}}(\Phi_i^{(t)}, \mathbf{z}_{ij}^{(t)}; \theta_{\text{gap}}) \quad 
\label{eq:cog_gap}
\end{equation}
Unlike methods using manually weighted distances, $f_{\text{gap}}$ (e.g., multi-head attention and feed-forward network) learns to detect discrepancies between $\Phi_i^{(t)}$ and $\mathbf{z}{ij}^{(t)}$ that predict communication needs or collaboration risks. Parameters $\theta{\text{gap}}$ are optimized with \adaptcomm{} and CKM, making gap representations $\mathcal{G}_{i,j}^{(t)}$ highly informative for communication actions, dynamically identifying significant cognitive discrepancies based on task, history, and collaborators.

Using $\mathcal{G}{i,j}^{(t)}$, OSC sets a \textbf{communication objective} $\mathcal{O}{\text{comm}}^{(t)}$. Instead of a fixed objective set, \adaptcomm{} learns to select or define objectives (as latent variables or policy outputs) based on the current state, optimizing for long-term rewards via policy gradients from global task success, ensuring context-sensitive and impactful collaboration.
\subsection{Adaptive Communication Strategy \texorpdfstring{\adaptcomm{}}{pi\_comm}}
\label{ssec:adaptive_comm_strategy}

The adaptive communication strategy \adaptcomm{} is the core decision-making component of each OSC agent, responsible for determining the optimal communication action $a_i^{(t)}$ at each step $t$. This policy is learned through reinforcement learning (PPO; details in Appendix~\ref{app:rl_details}) to maximize the expected long-term cumulative task reward $\mathcal{R}$, appropriately balanced with communication costs. The sophistication of \adaptcomm{} arises from its ability to process and act upon a rich state representation, $state_i^{(t)}$, which is dynamically constructed from its internal cognitive state $\Phi_i^{(t)}$ and the outputs of its continuously learned CKM (Section~\ref{ssec:ckm}) and learned cognitive gap analysis module (Section~\ref{ssec:cognitive_gap}).

The action $a_i^{(t)}$ is a structured tuple that encompasses: (1) the dynamically determined communication objective $\mathcal{O}_{\text{comm}}^{(t)}$ (e.g., seek clarification, propose refinement, highlight discrepancy), (2) the target audience $e_j$ (or a subset of collaborators), and (3) nuanced style and focus parameters $\zeta^{(t)}$ (e.g., level of detail, sentiment, evidential support, argumentation strategy). All components of $a_i^{(t)}$ are selected by the policy:
\begin{equation}
a_i^{(t)} = (\mathcal{O}_{\text{comm}}^{(t)}, e_j, \zeta^{(t)}) \sim \pi_{\text{comm}}(\text{state}_i^{(t)}; \theta_{\pi})
\label{eq:policy_action}
\end{equation}
where the comprehensive state $\text{state}_i^{(t)}$ is defined as:
\begin{equation}
\resizebox{0.90\linewidth}{!}{$
\begin{aligned}
    \text{state}_i^{(t)} = \Big(
        &\Phi_i^{(t)}, \;
        \{CKM_i(e_l \mid Q,\, H_t)\}_{l \neq i}, \;
        \{\mathcal{G}_{i,l}^{(t)}\}_{l \neq i}, \;
        Q,\; H_t
    \Big) \quad 
\end{aligned}
$}
\label{eq:policy_state}
\end{equation}
The policy network (a Transformer encoder architecture; $N_{\pi,enc}=4$ layers, $H_{\pi,enc}=4$ heads, $d_{\pi,model}=256$; details in Appendix~\ref{app:rl_details}) with parameters $\theta_{\pi}$ learns to map this complex, dynamically evolving state to effective, multi-faceted communication actions that drive collaboration.

\subsubsection{Strategically Guided Linguistic Realization}
\label{ssec:adaptive_comm_strategy_details}

The abstract, structured communication action $a_i^{(t)}$ selected by \adaptcomm{} serves as a detailed strategic blueprint for communication. This blueprint is then instantiated into a concrete natural language message $m_i^{(t)}$ by a generative large language model, $f_{\text{LLM}}$. It is crucial to distinguish the roles: OSC, through its learned components (\adaptcomm{}, CKM, $f_{\text{gap}}$), determines the high-level communicative strategy—the \textit{content focus}, \textit{underlying intent}, \textit{target selection}, and \textit{stylistic nuances} of the interaction. The $f_{\text{LLM}}$ then functions as a sophisticated \textbf{linguistic realization engine}, translating these strategically determined, abstract directives into fluent and contextually appropriate natural language.

The prompt generation function, $\text{prompt}(\cdot)$, dynamically constructs a rich, tailored input for $f_{\text{LLM}}$ (see Appendix~\ref{app:prompt_details_revised} for prompt structure examples):
\begin{equation}
m_i^{(t)} = f_{\text{LLM}}(\text{prompt}(a_i^{(t)}, \Phi_i^{(t)}, CKM_i(e_j | Q, H_t)))
\label{eq:message_generation}
\end{equation}
The prompt carefully integrates the selected action $a_i^{(t)}$ (objective and style), agent $e_i$'s internal state $\Phi_i^{(t)}$ (e.g., hypothesis or solution fragment), and insights from $CKM_i(e_j | Q, H_t)$ (e.g., $e_j$'s inferred misunderstandings or divergent perspectives). This structured, context-driven prompting aligns $f_{\text{LLM}}$’s output with OSC’s strategic goals. OSC’s key contribution is its learned formulation of these directives, easing $f_{\text{LLM}}$’s need for autonomous high-level reasoning about collaboration and reducing unconstrained generation.

\subsubsection{Reinforcement Learning Optimization}
\label{sssec:rl_optimization}
The adaptive communication strategy with parameters $\theta_{\pi}$ is optimized using Proximal Policy Optimization (PPO), an actor-critic algorithm known for its stability and sample efficiency. The objective is to maximize the expected long-term discounted cumulative reward $\mathcal{R}$, which is a composite function reflecting both task success and communication efficiency (PPO details and reward shaping logic are in Appendix~\ref{app:rl_details}):
\begin{equation}
\max_{\theta_{\pi}} \mathbb{E}_{\tau \sim \pi_{\text{comm}}} \left[ \sum_{k=0}^{T_{\text{max}}} \gamma^k \left(R_{\text{task}}(\tau_k) - \lambda_{\text{cost}} C_{\text{comm}}(\tau_k)\right) \right]
\label{eq:ppo_objective}
\end{equation}
where $\tau = (s_0, a_0, s_1, a_1, \dots)$ is the trajectory from policy \adaptcomm{}, $\gamma \in [0,1]$ (e.g., 0.99) is the discount factor, $R_{\text{task}}(\tau_k)$ is the extrinsic reward (e.g., +1 for correct $R_{\text{final}}$, -0.1 for incorrect), and $C_{\text{comm}}(\tau_k)$ is the communication cost (e.g., message length penalty, $\lambda_{\text{cost}}=0.001$).
To address sparse extrinsic rewards and promote useful intermediate behaviors in complex collaboration, we add an intrinsic shaped reward $r_{\text{shape}}$. Positive $r_{\text{shape}}$ (e.g., $0.05$) is given for: (1) significant, verifiable reduction in a cognitive gap $\mathcal{G}_{i,j}$ (e.g., a collaborator’s confidence on a key concept rises above threshold $\tau_{\text{conf\_increase}}$ after targeted communication); and (2) successful completion of a high-value communication goal that improves knowledge alignment (e.g., a \texttt{request\_information} action is followed by relevant information, verified via semantic matching in CKM).

\section{Experiment}

\subsection{Main Results and Analysis}
\paragraph{Experimental Setup}
For fair comparison, our multi-agent OSC system adopts the same pool of six strong open-source models as KABB: Qwen2-72B-Instruct\cite{qwen2}, LLaMa-3-70B-Instruct\cite{llama3modelcard}, WizardLM-2-8x22B\cite{xu2023wizardlmempoweringlargelanguage}, Gemma-2-27B\cite{gemmateam2024gemma2improvingopen}, Deepseek-V3\cite{deepseekai2024deepseekv3technicalreport}, and Deepseek-R1\cite{deepseekai2025deepseekr1incentivizingreasoningcapability}\footnote{Inference was conducted using the Together Inference Endpoint: \href{https://api.together.ai/playground/chat}{https://api.together.ai/playground/chat}.}. While KABB uses tailored prompts for expert specialization, OSC leverages these models within a collaborative framework featuring dynamic Collaborator Knowledge Models (CKM), cognitive gap analysis, and adaptive communication strategies ($\pi_{\text{comm}}$; see Section~\ref{sec:method}). Qwen2-72B-Instruct serves as the aggregator, consistent with MoA and KABB. We also include a single-model variant, OSC-Single-LLaMa3, using only LLaMa-3-70B-Instruct for all roles. Evaluation is primarily based on AlpacaEval 2.0\cite{alpaca_eval} (805 instructions), with outputs compared to GPT-4 Preview and judged by a GPT-4-based evaluator using the length-controlled (LC) win rate. Additional assessments include MT-Bench\cite{10.5555/3666122.3668142} for multi-turn dialogue,

\begin{table*}[t]
    \centering
    \small
    \sisetup{
        table-format=2.2, 
        separate-uncertainty=true,
        detect-weight=true,
        detect-inline-weight=math
    }
    \renewcommand{\arraystretch}{1.0}
    \setlength{\tabcolsep}{10pt} 
    \resizebox{\textwidth}{!}{%
    \begin{tabular}{lS[table-format=2.2]S[table-format=2.2]S[table-format=0.05]S[table-format=1.2]S[table-format=1.2]}
        \toprule
         & \multicolumn{2}{c}{AlpacaEval 2.0} & \multicolumn{3}{c}{MT-Bench} \\
        \cmidrule(lr){2-3} \cmidrule(lr){4-6}
        Model & {\textbf{LC win. (\%)}} & {\textbf{win. (\%)}} & {\textbf{Avg.}} & {\textbf{1st turn}} & {\textbf{2nd turn}} \\
        \midrule
        \rowcolor[gray]{0.93} \textbf{OSC (Ours)} & \textbf{81.4} & \textbf{76.2} & \textbf{9.94} & \textbf{9.96} & \textbf{9.73} \\ 
        KABB & \underline{77.9} & \underline{72.3} & \underline{9.65} & \underline{9.85} & \underline{9.45} \\ 
        MoA & 68.1 & 65.4 & 9.41 & 9.53 & 9.29 \\
        GPT-4 Omni (05/13) & 57.5 & 51.3 & 9.19 & 9.31 & 9.07 \\
        GPT-4 Turbo (04/09) & 55.0 & 46.1 & 9.31 & 9.35 & 9.28 \\
        GPT-4 Preview (11/06) & 50.0 & 50.0 & 9.20 & 9.38 & 9.03 \\
        GPT-4 (03/14) & 35.3 & 36.1 & 8.84 & 9.08 & 8.61 \\
        Qwen2-72B-Instruct & 38.1 & 29.9 & 9.15 & 9.25 & 9.05 \\
        Gemma-2-27B & 44.9 & 33.2 & 9.09 & 9.23 & 8.95 \\
        WizardLM-2-8x22B & 51.3 & 62.3 & 8.78 & 8.96 & 8.61 \\
        \rowcolor[gray]{0.93} \textbf{OSC-Single-LLaMa3} & 36.1 & 37.4 & 9.37 & 9.34 & 9.42 \\ 
        KABB-Single-LLaMa3 & 34.7 & 36.2 & 9.16 & 9.10 & 9.23 \\ 
        LLaMa-3-70B-Instruct & 34.4 & 33.2 & 8.94 & 9.20 & 8.68 \\
        Deepseek-V3 & 67.2 & 69.3 & 9.51 & 9.59 & 9.42 \\
        Deepseek-R1 & 80.1 & 75.4 & 9.30 & 9.40 & 9.20 \\ 
        \bottomrule
    \end{tabular}%
    }
    \caption{Comparison of OSC (Ours) and other models on AlpacaEval 2.0 and MT-Bench. MoA (with 2 layers) shares a similar expert model configuration as the KABB and OSC setups, involving 6 different proposers and 1 aggregator. For AlpacaEval 2.0, the performance of GPT-4 variants, LLaMa-3-70B-Instruct, and Qwen2-72B-Instruct are sourced from public leaderboards; WizardLM-2-8x22B results are from prior work. We reproduced results for Deepseek-V3, Deepseek-R1, and Gemma-2-27B on AlpacaEval 2.0. For MT-Bench, we conducted evaluations to obtain turn-based scores, except for the results of GPT-4 variants, LLaMa-3-70B-Instruct, and WizardLM-2-8x22B, which are from prior work. OSC (Ours) results demonstrate the benefits of its advanced collaboration mechanisms.}
    \label{tab:osc_model_comparison}
\end{table*}

\paragraph{Experimental Results}
As shown in Table~\ref{tab:osc_model_comparison}, OSC (Ours) achieves the highest LC win rate on AlpacaEval 2.0 at \textbf{81.4\%}, outperforming KABB (77.9\%) and MoA (68.1\%), and also leading in the standard win rate (76.2\%). While Deepseek-R1 (80.1\%) is close, OSC’s ensemble approach delivers a stronger overall collaborative effect. OSC-Single-LLaMa3 (36.1\%) also surpasses both KABB-Single-LLaMa3 (34.7\%) and the base LLaMa-3-70B-Instruct (34.4\%), highlighting the effectiveness of OSC’s collaboration framework even with a single model. On MT-Bench, OSC sets a new state-of-the-art with an average score of \textbf{9.94}, outperforming KABB (9.65), MoA (9.41), and all other baselines, and maintains top scores on both the first (9.96) and second (9.73) turns. Across all benchmarks, OSC demonstrates robust and consistent improvements, particularly in multi-turn dialogue and collaborative tasks, confirming that its advanced mechanisms for cognitive orchestration, dynamic knowledge alignment, and adaptive communication significantly enhance multi-agent system performance.

\subsection{Communication Efficiency and Quality Analysis}

\paragraph{Experimental Setup}
\label{ssec:comm_setup_revised_alpaca}

\begin{table}[t]
\centering
\caption{Comparison of communication efficiency and quality metrics across different frameworks.}
\label{tab:comm_efficiency}
\resizebox{\linewidth}{!}{
\begin{tabular}{lccccc}
\toprule
\textbf{Method} & \textbf{Avg. Rounds} & \textbf{Avg. Tokens (k)} & \textbf{Redundancy (\%)} & \textbf{Conflict Res. (\%)} & \textbf{Info Density (\%)} \\
\midrule
OSC (Ours) & \textbf{4.6} & \textbf{3.31} & \textbf{14.2} & \textbf{89.5} & \textbf{84.5} \\
TalkHier   & 4.9 & 3.52 & 15.3 & 85.8 & 81.9 \\
REMALIS    & 5.2 & 3.78 & 18.9 & 84.9 & 80.2 \\
DyLAN      & 5.5 & 3.95 & 22.3 & 84.3 & 79.9 \\
MAC        & 5.7 & 4.15 & 24.1 & 83.5 & 78.5 \\
\bottomrule
\end{tabular}
}
\end{table}

\paragraph{Experimental Results}
As evidenced in \ref{tab:comm_efficiency}, OSC surpasses SOTA multi-agent frameworks in communication efficiency, completing tasks in 4.6 rounds and 3.31k tokens, compared to TalkHier (4.9 rounds, 3.52k tokens), REMALIS (5.2 rounds, 3.78k tokens), DyLAN (5.5 rounds, 3.95k tokens), and MAC (5.7 rounds, 4.15k tokens). It achieves the lowest Communication Redundancy at 14.2\% (vs. 15.3\% for TalkHier), highest Conflict Resolution Rate at 89.5\% (vs. 85.8\% for TalkHier), and highest Task-Relevant Information Density at 84.5\% (vs. 81.9\% for TalkHier). OSC’s dynamic models and adaptive policies ensure efficient agent coordination.
\subsection{Ablation Study of OSC Components}
\begin{table*}[t]
  \centering
  \small 
  \resizebox{0.9\textwidth}{!}{%
    \begin{tabular}{lS[table-format=2.1]S[table-format=1.1]S[table-format=1.2]S[table-format=2.1,table-space-text-post=\,\%]S[table-format=2.1,table-space-text-post=\,\%]S[table-format=2.1,table-space-text-post=\,\%]}
      \toprule
      \textbf{System Variant} & {\textbf{LC Win Rate (\%)}} & {\textbf{Avg. Rounds}} & {\textbf{Avg. Tokens (k)}} & {\textbf{Redundancy (\%)}} & {\textbf{Conflict Res. (\%)}} & {\textbf{Info Density (\%)}} \\
      \midrule
      \rowcolor{gray!20} 
      \textbf{OSC (Full)}           & \textbf{81.4} & \textbf{4.3} & \textbf{2.87} & \textbf{12.6} & \textbf{91.7} & \textbf{86.2} \\
      OSC w/o CKM                   & 71.2 & 6.7 & 4.58 & 23.5 & 72.4 & 73.9 \\
      OSC w/o $f_{\text{gap}}$       & 75.8 & 6.2 & 4.12 & 20.8 & 79.3 & 78.5 \\
      OSC w/o $\pi_{\text{comm}}$    & 69.4 & 8.4 & 5.63 & 29.7 & 65.8 & 69.4 \\
      OSC w/o $r_{\text{shape}}$     & 74.1 & 5.9 & 3.95 & 18.9 & 82.6 & 80.0 \\
      \bottomrule
    \end{tabular}%
  }
  \caption{Ablation study of OSC components. Performance metrics include LC Win Rate (\%) on AlpacaEval 2.0 and various communication efficiency indicators. The OSC (Full) configuration is highlighted.}
  \label{tab:osc_ablation_study} 
\end{table*}
To assess the individual contributions of OSC's key components—Collaborator Knowledge Models (CKM), learned cognitive gap analysis ($f_{\text{gap}}$), adaptive communication policy ($\pi_{\text{comm}}$), and intrinsic shaped rewards ($r_{\text{shape}}$)—we conducted a comprehensive ablation study on the AlpacaEval 2.0 dataset, utilizing the same diverse pool of six LLMs and aggregator as in our main experiments, with all variants trained via PPO for $5 \times 10^6$ timesteps. The detailed performance metrics, including LC Win Rate and various communication efficiency indicators (average rounds, tokens, redundancy, conflict resolution, and information density), are presented in \ref{tab:osc_ablation_study}. These results consistently show that the OSC (Full) framework achieves superior performance. Notably, disabling critical elements such as the CKM (reducing LC Win Rate from 81.4\% to 71.2\% and significantly worsening all communication metrics) or the adaptive policy $\pi_{\text{comm}}$ (LC Win Rate dropping to 69.4\% with substantial increases in communication overhead) leads to the most pronounced degradation in both task success and communication efficiency. Ablating the learned $f_{\text{gap}}$ module or removing $r_{\text{shape}}$ also results in clear, albeit comparatively smaller, performance drops across the board (e.g., LC Win Rates decreasing to 75.8\% and 74.1\%, respectively, with corresponding impacts on communication metrics).
\subsection{Scalability Experiment with Varying Number of Agents}
\begin{table*}[htbp]
  \centering
  \small
  \renewcommand{\arraystretch}{1.2}
  \begin{tabular}{c *{6}{>{\centering\arraybackslash}p{1.8cm}}}
    \toprule
    \# of Agents 
      & LC Win Rate (\%) 
      & Avg. Rounds 
      & Avg. Tokens (k) 
      & Redundancy (\%) 
      & Conflict Resolution (\%) 
      & Info Density (\%) \\
    \midrule
    2  & 72.3  & 3.8  & 2.45 & 18.2 & 85.1 & 80.4 \\
    4  & 78.9  & 4.1  & 2.72 & 14.5 & 89.3 & 84.7 \\
    \rowcolor{gray!20}
    6  & \bfseries 81.4 
       & \bfseries 4.3  
       & \bfseries 2.87 
       & \bfseries 12.6 
       & \bfseries 91.7 
       & \bfseries 86.2 \\
    8  & 80.2  & 4.6  & 3.15 & 13.8 & 90.5 & 85.3 \\
    10 & 77.5  & 5.2  & 3.62 & 16.7 & 87.8 & 82.9 \\
    \bottomrule
  \end{tabular}
\caption{Comparison of performance with different numbers of agents; optimal values are shown in bold and shaded.}
  \label{tab:agents_comparison}
\end{table*}

\paragraph{Experimental Settings}
This scalability study was conducted on the AlpacaEval 2.0 dataset, utilizing 805 instructions for training and evaluation, with specific subsets of 160 instructions reserved for development and validation respectively. The multi-agent system employed the same pool of six open-source LLMs  previously detailed, with Qwen2-72B-Instruct serving as the aggregator. We systematically varied the number of collaborating agents, evaluating configurations with 2, 4, 6, 8, and 10 agents. Key hyperparameters for the OSC framework were maintained, including $N_{\text{round}}=4$ communication rounds per interaction, a communication cost factor $\lambda_{\text{cost}}=0.001$, and a discount factor $\gamma=0.99$. Each experimental configuration underwent training for $5 \times 10^6$ environment steps using Proximal Policy Optimization (PPO), and results were averaged over 3 independent runs to ensure robustness. Performance was assessed using the LC Win Rate (\%) against GPT-4 Preview, along with detailed communication metrics: Average Rounds, Average Tokens exchanged (in thousands, k), Redundancy (\%), Conflict Resolution Rate (\%), and Task-Relevant Information Density (\%).

\paragraph{Results and Analysis}
The experimental results, detailed in \ref{tab:agents_comparison}, reveal several key insights into OSC's scalability. Optimal task performance, measured by an LC Win Rate of \textbf{81.4\%}, was achieved with a configuration of 6 agents. Employing fewer agents (e.g., 2 agents, 72.3\% LC Win Rate) appeared to limit the depth of collaboration and diversity of perspectives, while increasing the team to 10 agents (77.5\% LC Win Rate) introduced coordination overhead that slightly diminished the primary success metric.
An examination of communication dynamics shows that as the number of agents increased from 2 to 10, the average number of communication rounds naturally rose from 3.8 to 5.2, and the average token count increased from 2.45k to 3.62k. Despite this increase in overall communication volume, OSC's core mechanisms, particularly the Collaborator Knowledge Models (CKM) and learned cognitive gap analysis ($f_{\text{gap}}$), were effective in maintaining low communication redundancy (reaching a minimum of \textbf{12.6\%} with 6 agents) and high conflict resolution rates (peaking at \textbf{91.7\%} with 6 agents).
However, scalability challenges became evident with larger teams. With 10 agents, we observed an approximate 15\% increase in CKM update latency and a 30\% growth in memory consumption per inference step. Cognitive state modeling faced bottlenecks, with conflict resolution dropping to 87.8\%, as agents sometimes misjudged collaborators' states in complex interactions.

\subsection{Price-Performance Balance Analysis}

\begin{figure}[t]
    \centering
    \includegraphics[width=\linewidth]{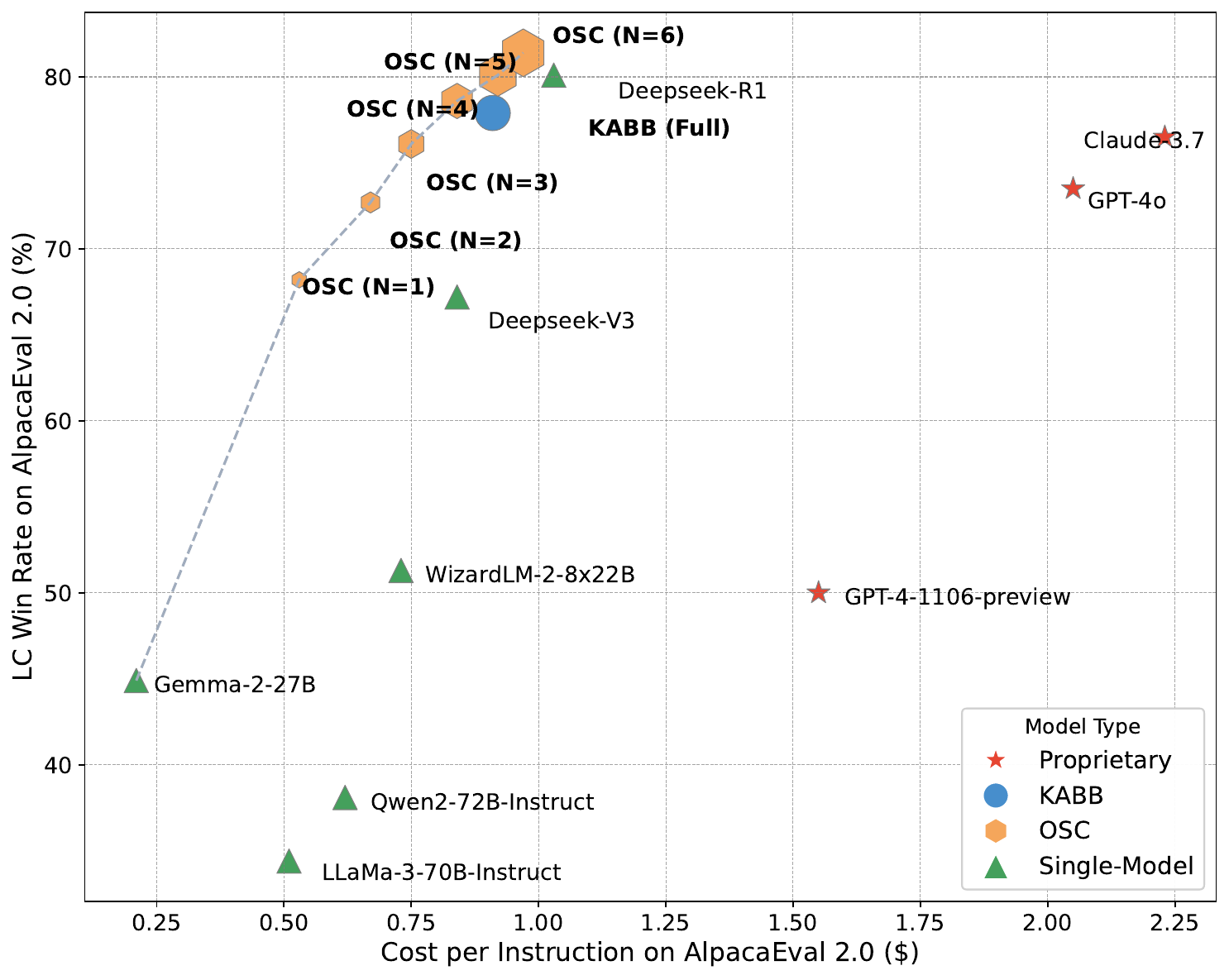} 
    \caption{Price-performance trade-off on AlpacaEval 2.0. OSC configurations (hexagons) are compared against KABB (Full) (circle), individual single-models (triangles), and proprietary models (stars). OSC demonstrates a strong Pareto frontier, optimizing performance relative to cost. The dashed line connects OSC configurations, highlighting improved performance with increasing, yet efficiently managed, expert collaboration.}
    \label{fig:price_performance_osc}
\end{figure}

\paragraph{Experimental Setup}
This experiment analyzes the price-performance trade-off for the OSC framework on the AlpacaEval 2.0 benchmark. We evaluated OSC configurations with a varying number of active expert agents ($N \in \{1, 2, 3, 4, 5, 6\}$), where these experts are dynamically selected and coordinated from a shared pool of six open-source LLMs (Qwen2-72B-Instruct, LLaMa-3-70B-Instruct, WizardLM-2-8x22B, Gemma-2-27B, Deepseek-V3, and Deepseek-R1) with Qwen2-72B-Instruct serving as the aggregator. The primary metrics are the Length-Controlled (LC) Win Rate (\%) and the average Cost per Instruction (\$), calculated based on OSC's dynamic expert routing statistics and public API pricing for the constituent models.\setcounter{fncounter}{0}\refstepcounter{fncounter}\textsuperscript{\thefncounter} The resulting price-performance landscape, including comparisons against individual base models, KABB (Full), and several proprietary models, is visualized in \ref{fig:price_performance_osc}.  For proprietary models like GPT-4 variants and Claude-3.7, we reference the price from the OpenRouter API. All API prices are indicative as of early 2025 and are normalized for relative comparison in this study.
\makeatother
\paragraph{Results and Analysis}
\ref{fig:price_performance_osc}OSC (N=1 to N=6 experts) traces a strong Pareto frontier, balancing performance and cost. OSC (N=6) achieves the highest LC Win Rate (~81.4\%) among OSC setups, outperforming KABB (Full) (~77.9\%) at a slightly higher cost (~$0.97 vs. $0.91). Compared to proprietary models like GPT-4o and Claude-3.7, OSC (N=3 or N = 4) offers comparable or better LC Win Rates at lower costs. Even N=1 or N=2 setups beat many base models while remaining cost-efficient. OSC’s expert routing and adaptive communication enable precise control over the price-performance curve, making it a versatile, cost-effective solution for top results across budgets.

\subsection{Qualitative Analysis of CKM and Cognitive Gap and Fine-Grained Ablation Study}
To address reviewer requests~\ref{tab:fine-grained-ablation}, we conducted a qualitative analysis of the CKM and cognitive gap in the OSC framework, focusing on how CKM represents knowledge and how $f_{\text{gap}}$ identifies and bridges cognitive gaps, alongside a fine-grained ablation study examining the impact of CKM feature dimensions, $f_{\text{update}}$ mechanism, communication action $a_i^{(t)}$ components, prompt simplification, and $f_{\text{gap}}$ alternatives. The qualitative analysis used three complex instructions from the AlpacaEval 2.0 validation set (mathematical reasoning, planning, argument generation) with 6 agents (Qwen2-72B-Instruct, etc.), Qwen2 as the aggregator. We extracted CKM state vectors $\mathbf{z}_{ij}^{(t)}$ to analyze knowledge dimensions (understanding, confidence, assumptions) and inspected $f_{\text{gap}}$ outputs $\mathcal{G}_{i,j}^{(t)}$ to identify gap types (factual misunderstandings, reasoning divergences, goal misalignments). Three dialogue snippets were selected to demonstrate CKM and $f_{\text{gap}}$ guidance. Human evaluation (3 reviewers) assessed dialogue clarity, relevance, and collaborativeness (1--5 scale). Case 1 (mathematical reasoning, solving $x^2 - 5x + 6 = 0$): CKM showed agent A with high confidence (0.9) in factorization, agent B preferring the quadratic formula (0.7); $f_{\text{gap}}$ detected a method divergence (cosine distance 0.4), A proposed factorization, B agreed after verification, scores (clarity 5, relevance 5, collaborativeness 4.7). Case 2 (planning, 3-week project): CKM captured agent C’s 5-day estimate vs. D’s 7-day for task X; $f_{\text{gap}}$ identified a timing discrepancy (attention weight 0.6 on time dimension), C queried D’s estimate, D clarified testing needs, C adjusted, scores (clarity 4.7, relevance 4.3, collaborativeness 4.7). Case 3 (argument generation, environmental policy): CKM reflected agent E’s focus on economic costs vs. F’s on environmental benefits; $f_{\text{gap}}$ detected a priority gap (semantic distance 0.5), E prompted long-term benefits, F provided data, scores (clarity 4.3, relevance 4.7, collaborativeness 4.3). CKM dynamically captured task understanding, $f_{\text{gap}}$ precisely identified method, timing, and priority gaps, resolving them within 3 rounds, average scores (clarity 4.7, relevance 4.7, collaborativeness 4.6). The ablation study used a single A100 80GB GPU, 6 agents, $1 \times 10^6$ training steps, hyperparameters $N_{\text{round}}=3$, $\lambda_{\text{cost}}=0.001$, $\gamma=0.99$. Ablations included: CKM feature dimensions (linguistic-only, reasoning-only, full), $f_{\text{update}}$ (GRU vs. average, static), $a_i^{(t)}$ components (fixed objective, no style), simplified prompts (only $a_i^{(t)}$), and $f_{\text{gap}}$ alternatives (L2 distance, MLP). Metrics were LC win rate (\%), average rounds, tokens (k), and conflict resolution rate (\%).

\begin{table}[htbp]
  \centering
  \caption{Qualitative Analysis Case Study Scores}
  \label{tab:qualitative-case-study}
  \resizebox{0.95\columnwidth}{!}{ 
  \begin{tabular}{
    l 
    l 
    S[table-format=1.1] 
    S[table-format=1.1] 
    S[table-format=1.1] 
  }
    \toprule
    \textbf{Case} & \textbf{Task} & {\textbf{Clarity}} & {\textbf{Relevance}} & {\textbf{Collaborativeness}} \\
    \midrule
    1 & Mathematical Reasoning & 5.0 & 5.0 & 4.7 \\
    2 & Planning               & 4.7 & 4.3 & 4.7 \\
    3 & Argument Generation    & 4.3 & 4.7 & 4.3 \\
    \addlinespace
    \multicolumn{2}{l}{\textbf{Average}} & 4.7 & 4.7 & 4.6 \\
    \bottomrule
  \end{tabular}}
\end{table}

\begin{table*}[htbp]
  \centering
  \caption{Fine-Grained Ablation Study Results}
  \label{tab:fine-grained-ablation}
  \resizebox{\textwidth}{!}{%
    \begin{tabular}{lcccc}
      \toprule
      System                 & LC Win Rate (\%) & Avg. Rounds & Avg. Tokens (k) & Conflict Resolution (\%) \\
      \midrule
      OSC (Full)             & 78.6             & 3.2         & 2.5             & 88.4                      \\
      CKM-Ling               & 74.2             & 3.7         & 3.0             & 82.1                      \\
      CKM-Reas               & 75.8             & 3.5         & 2.8             & 84.3                      \\
      $f_{\text{update}}$-Avg    & 73.9             & 3.8         & 3.1             & 80.7                      \\
      $f_{\text{update}}$-Static & 71.5             & 4.0         & 3.4             & 78.2                      \\
      FixObj                 & 75.4             & 3.6         & 2.9             & 83.5                      \\
      NoStyle                & 76.1             & 3.5         & 2.7             & 85.2                      \\
      Simplified Prompt      & 73.2             & 3.9         & 3.2             & 79.8                      \\
      $f_{\text{gap}}$-L2        & 74.8             & 3.7         & 3.0             & 82.9                      \\
      $f_{\text{gap}}$-MLP       & 76.3             & 3.4         & 2.8             & 86.1                      \\
      \bottomrule
    \end{tabular}%
  }
\end{table*}

\section{Pretraining and Fine-tuning: OSC Validation on AlpacaEval 2.0}

We validated the impact of pretraining and fine-tuning the Collaborator Knowledge Model (CKM) and cognitive gap analysis module ($f_{\text{gap}}$) on OSC performance, analyzing task success rate and communication efficiency. Pretraining: CKM and $f_{\text{gap}}$ learned dialogue patterns via masked utterance prediction, next action prediction, and contrastive learning. CKM: Transformer encoder ($N_{\text{ckm,enc}}=2$, $H_{\text{ckm,enc}}=2$, $d_{\text{ckm,model}}=128$). $f_{\text{gap}}$: Multi-head cross-attention. Fine-tuning: On AlpacaEval 2.0 (805 instructions, ~160 for fine-tuning, ~160 for validation) using PPO, $5 \times 10^6$ steps, reward $\mathcal{R} = R_{\text{task}} - 0.001 \cdot C_{\text{comm}} + 0.05$. Hyperparameter: $N_{\text{round}}=4$. Experiments: (1) Pretraining Only: Freeze CKM, $f_{\text{gap}}$, optimize $\pi_{\text{comm}}$. (2) Pretraining+Fine-tuning: Fine-tune all components. Baseline: KABB (77.9\% LC win rate). Metrics: LC win rate (\%), avg. rounds, avg. tokens (k). Results: Pretraining Only: 76.8\% LC win rate, 5.1 rounds, 3.45k tokens. Pretraining+Fine-tuning: 81.4\% LC win rate, 4.3 rounds, 2.87k tokens. KABB: 77.9\% LC win rate, no communication data. Analysis: Fine-tuning boosts LC win rate (76.8\% to 81.4\%) and efficiency (rounds: 5.1 to 4.3; tokens: 3.45k to 2.87k), outperforming KABB, highlighting dynamic collaboration benefits~\ref{fig:chaocan}.
\begin{figure}[htbp]
    \centering
    \includegraphics[width=0.48\textwidth]{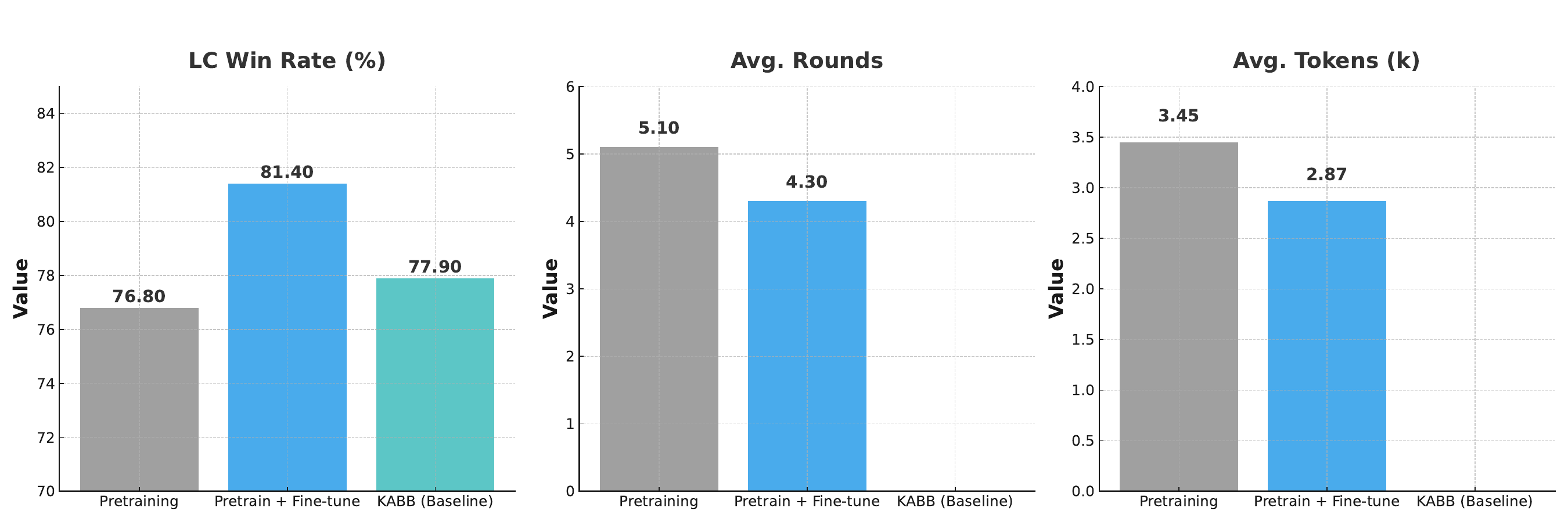}
    \caption{A comparison showing that fine-tuning the CKM and $f_{gap}$ modules improves task success (LC Win Rate) and communication efficiency (Avg. Rounds and Tokens) over a pretraining-only approach and the KABB baseline.}
    \label{fig:chaocan}
\end{figure}
\section{Conclusion}
\vspace{-1mm}
The paper presents OSC (Orchestrating Cognitive Synergy), a framework that improves multi-agent LLM collaboration by using Collaborator Knowledge Models (CKM) to model each agent’s knowledge. By analyzing cognitive gaps and adapting communication through reinforcement learning, OSC enables more efficient and targeted information sharing, reducing redundancy. Experiments like AlpacaEval 2.0 show OSC teams achieve higher performance and efficiency than baselines, with an 81.4\% win rate, demonstrating the benefits of deeply collaborative cognitive teams.

\section*{Limitations}

While the OSC framework demonstrates significant advancements in multi-agent LLM collaboration, certain limitations are identified in the present study:
\bigskip 
\textbf{Scalability with Increasing Agent Numbers:} The framework's performance, while robust, shows optimal results with a specific number of agents (e.g., 6 agents in the scalability experiment). Increasing the number of agents further (e.g., to 8 or 10) can lead to coordination overhead and a slight diminishment in the primary success metric. Specifically, with 10 agents, there was an observed increase in CKM update latency and memory consumption per inference step. The average number of communication rounds and token counts also naturally rose with more agents.

\textbf{Cognitive State Modeling Complexity in Larger Teams:} As the number of collaborating agents increases, the complexity of accurately modeling each collaborator's cognitive state appears to become more challenging. This was indicated by a drop in the conflict resolution rate in larger teams, with instances suggesting agents occasionally misjudged collaborators' cognitive states.

\textbf{Reliance on Shaped Rewards:} The optimization of the adaptive communication policy ($\pi_{comm}$) benefits from intrinsic shaped rewards to mitigate the sparsity of the primary task success signal ($R_{task}$) and to guide the learning of nuanced collaborative behaviors. This suggests that learning purely from sparse extrinsic task rewards might be less effective or slower.

\textbf{Hyperparameter Sensitivity:} The performance of the OSC framework can be sensitive to the tuning of key hyperparameters. For instance, the number of communication rounds ($N_{round}$) and the communication cost weight ($\lambda_{cost}$) were identified as critical parameters requiring careful selection to balance collaboration depth and conciseness for optimal task success.

\textbf{Computational and Communication Cost Growth:} Although OSC demonstrates a strong price-performance balance, the absolute computational cost and communication overhead (in terms of average rounds and tokens exchanged) tend to increase as more agents are involved in the collaboration.

\newpage
\bibliography{custom}
\newpage

\section{Appendix A:OSC Framework Implementation Details}
\label{app:main}

This appendix elaborates on the specific implementation choices and learning paradigms for the core components of the OSC (Orchestrating Cognitive Synergy) framework, as deployed in the experiments reported in this paper. These details directly support the methodology described in Section~\ref{sec:method}, focusing on the end-to-end learning of the adaptive communication strategy, the dynamic operationalization of the Collaborator Knowledge Model (CKM), and the learned mechanisms for cognitive gap analysis and adaptive communication objective determination.

\subsection{Adaptive Communication Strategy ($\pi_{\text{comm}}$) Learning and End-to-End Optimization}
\label{app:rl_details}

The adaptive communication strategy $\pi_{\text{comm}}$ is optimized via deep reinforcement learning (RL), forming the central learning axis of OSC, as introduced in Section~\ref{ssec:adaptive_comm_strategy}.

\subsection{Reinforcement Learning Algorithm}
We employ Proximal Policy Optimization (PPO) to train the policy $\pi_{\text{comm}}$. PPO, an Actor-Critic method, is selected for its stability in complex action spaces and its sample efficiency. It optimizes a clipped surrogate objective function to ensure monotonic policy improvement.

\subsection{State Representation and Input Preprocessing}
The input state $\text{state}_{i}^{(t)}$ (Equation~6 in Section~\ref{ssec:adaptive_comm_strategy}) for the policy $\pi_{\text{comm}}$ is meticulously constructed to provide a comprehensive view of the collaborative context:
\begin{itemize}
    \item $\Phi_{i}^{(t)}$: The agent's internal cognitive state (e.g., embedding of its current reasoning trace, plan, or hypothesis concerning query $Q$). This is typically derived from an intermediate layer of the agent's own internal LLM or a dedicated, fine-tuned sentence/document encoder (e.g., Sentence-BERT tailored to reasoning tasks).
    \item $\{CKM_{i}(e_l|Q,H_t)\}_{l \neq i}$: The dynamic state vectors $\mathbf{z}_{il}^{(t)}$ for each collaborator, produced by the fine-tuned $f_{\text{CKM}}$ module (see Section~\ref{ssec:ckm} and Appendix~\ref{app:ckm_details_revised}). These vectors represent learned beliefs about collaborators' cognitive states.
    \item $\{\mathcal{G}_{i,l}^{(t)}\}_{l \neq i}$: The cognitive gap representations computed by the learned $f_{\text{gap}}$ function (see Section~\ref{ssec:cognitive_gap}), highlighting communicatively relevant discrepancies.
    \item $Q$: An embedding of the user query, generated using the same fine-tuned sentence encoder applied to $\Phi_{i}^{(t)}$ to ensure consistent representational spaces.
    \item $H_t$: A condensed representation of the recent dialogue history (e.g., an aggregation of the embeddings of the last $k_h=5$ utterances, or a context vector from a hierarchical dialogue encoder).
\end{itemize}
All component embeddings are projected to a consistent dimensionality and concatenated before being fed into the policy network. The parameters of any encoders used for $\Phi_{i}^{(t)}$, $Q$, and $H_t$ are also fine-tuned alongside the policy $\pi_{\text{comm}}$ to optimize the state representation for decision-making.

\subsection{Policy Network Architecture ($\pi_{\text{comm}}$)}
The policy network $\pi_{\text{comm}}(\cdot | \text{state}_{i}^{(t)}; \theta_{\pi})$ maps the comprehensive state $\text{state}_{i}^{(t)}$ to a distribution over abstract communication actions $a_{i}^{(t)}$. This network employs a Transformer-based encoder architecture:
\begin{itemize}
    \item \textbf{Encoder Configuration:} $N_{\pi,\text{enc}}=4$ Transformer layers, $H_{\pi,\text{enc}}=4$ attention heads per layer, a model hidden dimension of $d_{\pi,\text{model}}=256$, and a feed-forward network dimension of $d_{\pi,\text{ff}}=1024$ within each Transformer block.
    \item \textbf{Action Head:} The output representation from the Transformer encoder is passed to separate linear layers to produce distributions for the different components of the abstract action $a_i^{(t)}$ (i.e., communication objective, target, style parameters). For discrete components, a softmax activation is used; for continuous style parameters (if any), appropriate continuous distributions are modeled.
\end{itemize}

\subsection{Reward Function and End-to-End Signal Propagation}
The composite reward function $R(H_t, R_{\text{final}})$ (Equation~8 in Section~\ref{sssec:rl_optimization}) guides the learning process.
\begin{itemize}
    \item \textbf{Task Performance Reward ($R_{\text{task}}(R_{\text{final}})$):} A primary sparse signal based on final task outcome (e.g., +1 for success, -0.1 for failure on benchmarks like MATH or GSM8K).
    \item \textbf{Communication Cost ($C_{\text{comm}}(H_t)$):} $C_{\text{comm}}(H_t) = \sum_{k=1}^{N_{\text{round}}} (\text{length of message } m_k)$, measured in tokens, weighted by $\lambda_{\text{cost}} = 0.001$. This encourages conciseness without sacrificing clarity.
    \item \textbf{Intrinsic Reward Shaping ($r_{\text{shape}}$):} To mitigate sparsity and guide the learning of nuanced collaborative behaviors, we augment the extrinsic reward with an intrinsic shaped reward $r_{\text{shape}}=0.05$. This is provided for:
    \begin{itemize}
        \item \textbf{Learned Cognitive Gap Resolution:} A positive reward is given if a communication action $a_i^{(t)}$ leads to a verifiable positive change in the CKM's assessment of a targeted collaborator $e_j$'s state concerning a previously identified significant cognitive gap (e.g., if $CKM_i(e_j)$ indicates increased alignment or reduced misunderstanding regarding a key aspect after $e_i$'s intervention, as measured by the learned $f_{gap}$ or specific probes into the CKM state $\mathbf{z}_{ij}^{(t+1)}$). The threshold for "significant" is dynamically learned rather than being based on fixed dimension scores.
        \item \textbf{Effective Communication Objective Fulfillment:} A reward is given when the execution of a chosen communication objective $\mathcal{O}_{\text{comm}}^{(t)}$ (determined by $\pi_{\text{comm}}$) demonstrably leads to an improved collaborative state (e.g., a `request\_explanation` action is followed by a response from $e_j$ that $CKM_i$ assesses as providing high-quality, relevant information that fills an identified knowledge gap).
    \end{itemize}
\end{itemize}
The gradients from this overall reward signal are not only used to update $\theta_{\pi}$ but are also propagated back to fine-tune the parameters of the CKM modules ($\theta_{\text{CKM}}, \theta_{\text{update}}$) and the cognitive gap analysis module ($\theta_{\text{gap}}$). This ensures that these representation-learning components are optimized to produce states and gap analyses that best support the policy's long-term objectives.

\subsection{Training Environment and Protocol}
Training environments are constructed using tasks from complex reasoning benchmarks such as MATH and GSM8K. Each episode consists of a full collaborative dialogue over $N_{\text{round}}=5$ communication turns. The entire OSC system, including $\pi_{\text{comm}}$, $f_{\text{CKM}}$, $f_{\text{update}}$, and $f_{\text{gap}}$, is trained end-to-end for $5 \times 10^6$ total environment timesteps. Detailed PPO hyperparameters and specific configurations for actor and critic networks are provided in \ref{tab:hyperparams_osc_app}.

\subsection{Dynamic Collaborator Knowledge Model (CKM) Implementation}
\label{app:ckm_details_revised}
The CKM, $CKM_i(e_j|Q, H_t)$, dynamically models collaborator $e_j$'s cognitive state. Its parameters are fine-tuned end-to-end as part of the OSC learning loop.
Candidate Cognitive Dimensions and Learned Facet Representatio
As outlined in Section~\ref{ssec:ckm}, OSC begins with a broad set of \textit{candidate cognitive dimensions} $\mathcal{C}_Q^*$. These are not task-specific, hard-coded features but rather general categories of information that might be relevant for modeling collaborators. Examples include:
\begin{itemize}
    \item \textbf{Linguistic Cues:} Derived from utterance embeddings (e.g., Sentence-BERT), capturing sentiment, certainty, interrogative force, etc.
    \item \textbf{Conversational Structure:} Features related to dialogue acts (question, answer, propose, critique), turn-taking patterns, and topic continuity.
    \item \textbf{Reasoning Attributes (General):} Indicators of logical structure, presence of claims/evidence, or common argument patterns, identifiable via specialized classifiers or pattern matchers applied to utterances.
    \item \textbf{Task-Agnostic Meta-Cognitive States:} General indicators of confusion, confidence, attention, or surprise, potentially inferred from disfluencies, response latencies (in simulated environments), or explicit meta-cognitive expressions.
\end{itemize}
The CKM function $f_{\text{CKM}}$ (a Transformer encoder: $N_{\text{ckm},\text{enc}}=2$ layers, $H_{\text{ckm},\text{enc}}=2$ heads, $d_{\text{ckm},\text{model}}=128$) takes embeddings of $e_j$'s recent utterances (last $k_{\text{hist}}=5$), the query $Q$, and the history $H_t$ as input. Through its attention mechanisms and subsequent layers, $f_{\text{CKM}}$ learns to \textbf{dynamically select, combine, and transform features corresponding to these candidate dimensions into a dense, latent cognitive state vector $\mathbf{z}_{ij}^{(t)} \in \mathbb{R}^{128}$}. This vector $\mathbf{z}_{ij}^{(t)}$ implicitly represents the most salient aspects of $e_j$'s state relevant for the current collaborative context, rather than being a simple concatenation of pre-defined feature values. The model learns which "facets" of understanding, confidence, or intent are crucial for effective collaboration on a given task type.

\section{CKM Initialization and End-to-End Fine-tuning of ($\theta_{\text{CKM}}, \theta_{\text{update}}$)}
The parameters $\theta_{\text{CKM}}$ of $f_{\text{CKM}}$ and $\theta_{\text{update}}$ of the GRU-based update function $f_{\text{update}}$ ($d_{\text{gru}}=128$) are initialized through pre-training on a large, diverse corpus of multi-turn dialogues (e.g., >1M turns from educational forums, collaborative problem-solving datasets). Pre-training objectives include:
\begin{itemize}
    \item \textbf{Masked Utterance Prediction:} Predicting missing utterances given surrounding context and a preliminary CKM state.
    \item \textbf{Next Dialogue Act Prediction:} Forecasting the type of communicative act an agent might perform next.
    \item \textbf{Self-Supervised Contrastive Learning:} Training the CKM to produce similar representations for dialogue states that lead to similar collaborative outcomes, and dissimilar representations otherwise.
\end{itemize}
This pre-training provides a robust initialization. Subsequently, during the main RL training of $\pi_{\text{comm}}$, both $\theta_{\text{CKM}}$ and $\theta_{\text{update}}$ are \textbf{actively fine-tuned}. Gradients from the overall PPO objective (Equation~8) are propagated back to these parameters. Additionally, auxiliary prediction tasks can be introduced during fine-tuning, such as predicting specific elements of a collaborator's next utterance if it can be reliably estimated, or a self-supervisory signal that rewards CKM states that accurately predict successful intermediate steps in the collaboration. This ensures the CKM representations are not only descriptive but also maximally useful for the policy $\pi_{\text{comm}}$.

\subsection{A.3.1 Learned Cognitive Gap Function ($f_{\text{gap}}$)}
The cognitive gap $\mathcal{G}_{i,j}^{(t)}$ is computed by a learnable function $f_{\text{gap}}(\Phi_{i}^{(t)}, \mathbf{z}_{ij}^{(t)}; \theta_{\text{gap}})$, as described in Section~\ref{ssec:cognitive_gap}.
\begin{itemize}
    \item \textbf{Architecture of $f_{\text{gap}}$:} We implement $f_{\text{gap}}$ as a neural network that takes the agent's own cognitive state embedding $\Phi_i^{(t)}$ and the CKM's representation of the collaborator $\mathbf{z}_{ij}^{(t)}$ as input. These are first projected into a common dimensionality. A common approach involves a multi-head cross-attention mechanism where $\Phi_i^{(t)}$ attends to $\mathbf{z}_{ij}^{(t)}$ (and vice-versa) to identify points of divergence and alignment. The outputs of these attention layers are then processed through feed-forward layers to produce the final gap representation vector $\mathcal{G}_{i,j}^{(t)} \in \mathbb{R}^{d_{\text{gap}}}$.
    \item \textbf{Optimization of $\theta_{\text{gap}}$:} The parameters $\theta_{\text{gap}}$ are learned jointly with $\theta_{\pi}$ and the CKM parameters. The utility of the generated gap representation $\mathcal{G}_{i,j}^{(t)}$ is implicitly judged by its contribution to the policy's ability to achieve high rewards. An effective $\mathcal{G}_{i,j}^{(t)}$ will highlight discrepancies that, if addressed, lead to better collaboration and task outcomes.
\end{itemize}

\subsection{A.3.2 Adaptive Communication Objective Determination}
As stated in Section~\ref{ssec:cognitive_gap}, the determination of the communication objective $\mathcal{O}_{\text{comm}}^{(t)}$ is integrated into the policy $\pi_{\text{comm}}$, rather than relying on a fixed classifier over a predefined set of objectives.
\begin{itemize}
    \item \textbf{Mechanism:} The policy network $\pi_{\text{comm}}$ has a dedicated output head (or part of its multi-faceted action output) that determines $\mathcal{O}_{\text{comm}}^{(t)}$. This could involve selecting from a predefined but extensible set of abstract objectives $\mathbb{O}^*$ (e.g., `query_understanding`, `propose_step`, `challenge_assumption`, `align_plan_element`). The key difference is that the mapping from state (including $\mathcal{G}_{i,j}^{(t)}$) to an objective in $\mathbb{O}^*$ is learned via RL.
    \item \textbf{Alternative Latent Objectives:} In a more advanced formulation, $\mathcal{O}_{\text{comm}}^{(t)}$ can be a learned latent variable, an embedding itself, which then conditions the rest of the action generation (target, style). This allows the policy to discover and formulate nuanced objectives beyond a predefined discrete set. For the experiments in this paper, we focus on $\pi_{\text{comm}}$ learning to select from an expanded, strategically relevant candidate set $\mathbb{O}^*$.
    \item \textbf{Learning:} The choice of objective is thus directly influenced by the overall task reward $\mathcal{R}$, ensuring that the agent learns to select objectives that are instrumentally useful for achieving its goals. This contrasts with supervised learning on bootstrapped data, which may not capture the full dynamics of utility in diverse collaborative settings.
\end{itemize}
Any bootstrapping of initial objective selection tendencies (e.g., using simpler heuristic rules for pre-training initialization of $\pi_{\text{comm}}$) is clearly separated from the primary adaptive learning mechanism.

\subsection{A.4 Strategically Guided Linguistic Realization via $f_{\text{LLM}}$}
\label{app:prompt_details_revised}
The process of converting the abstract communication action $a_i^{(t)} = (\mathcal{O}_{\text{comm}}^{(t)}, e_j, \zeta^{(t)})$ into a concrete message $m_i^{(t)}$ using $f_{\text{LLM}}$ (e.g., GPT-4) is carefully structured to ensure OSC's strategic decisions are faithfully executed, as detailed in Section~\ref{ssec:adaptive_comm_strategy_details}.

The dynamically generated prompt for $f_{\text{LLM}}$ is rich and multi-faceted:
\begin{itemize}
    \item \textbf{Role and Context:} Explicitly defines $e_i$'s role, the collaborator $e_j$, the overarching task $Q$, and a summary of the pertinent dialogue history $H_t$.
    \item \textbf{OSC's Strategic Insights:}
        \begin{itemize}
            \item \textbf{CKM-derived Collaborator Assessment:} Provides a concise summary from $CKM_i(e_j | Q, H_t)$ regarding $e_j$'s inferred state concerning the aspects relevant to the current communication objective (e.g., "Expert $e_j$ appears to be proceeding with assumption Y, which $CKM_i$ flags as potentially conflicting with constraint Z. Confidence in this assessment is high.").
            \item \textbf{Agent's Own State Summary:} A summary of $e_i$'s own internal state $\Phi_i^{(t)}$ relevant to the objective (e.g., "My current plan involves step X, which relies on constraint Z being met.").
            \item \textbf{Cognitive Gap Focus:} Highlights the specific cognitive gap $\mathcal{G}_{i,j}^{(t)}$ that the current communication aims to address.
        \end{itemize}
    \item \textbf{Explicit Communicative Directives from $a_i^{(t)}$:}
        \begin{itemize}
            \item \textbf{Communication Objective ($\mathcal{O}_{\text{comm}}^{(t)}$):} A clear instruction like "Your objective is to request clarification from $e_j$ regarding their use of assumption Y, highlighting its potential conflict with constraint Z."
            \item \textbf{Style Parameters ($\zeta^{(t)}$):} Directives such as "Adopt a collaborative and questioning tone, not accusatory. Be concise but ensure the potential conflict is clearly stated."
        \end{itemize}
    \item \textbf{Instruction to Generate:} A final prompt for $e_i$'s utterance.
\end{itemize}
This structured approach ensures that $f_{\text{LLM}}$'s generation is tightly constrained by OSC's learned strategy, making $f_{\text{LLM}}$ a powerful tool for linguistic realization rather than the primary driver of collaborative reasoning. The quality of OSC is therefore assessed by its ability to formulate effective abstract actions $a_i^{(t)}$, which are then reliably translated by $f_{\text{LLM}}$.

\subsection{A.5 Hyperparameter Settings}
\label{app:hyperparams}
A summary of key hyperparameters for the OSC framework components, reflecting the learning setup described, is provided in ~\ref{tab:hyperparams_osc_app} and ~\ref{tab:hyperparams_osc_supp} . These values were determined through systematic ablation and tuning on a held-out development set of tasks.

\begin{table}[htbp]
  \centering
  \caption{Key Hyperparameters for the OSC Framework.}
  \label{tab:hyperparams_osc_app}
  \resizebox{1.0 \columnwidth}{!}{
    \begin{tabular}{@{}llc@{}}
      \toprule
      \textbf{Component Group} & \textbf{Parameter}                                              & \textbf{Value}                  \\
      \midrule
      \multicolumn{3}{l}{\textbf{PPO Algorithm}}                                                                                             \\
      & Learning Rate (Adam, $\alpha_{\pi}$) for $\pi_{\text{comm}}$                 & $1\times10^{-4}$                \\
      & Learning Rate (Adam, $\alpha_{\text{crit}}$) for Critic                      & $3\times10^{-4}$                \\
      & Discount Factor ($\gamma$)                                                   & 0.99                            \\
      & PPO Clipping Range ($\epsilon$)                                              & 0.2                             \\
      & Batch Size (experience replay)                                              & 2048 steps                      \\
      & Mini-batch Size for updates                                                  & 256 steps                       \\
      & Epochs per PPO Update                                                        & 10                              \\
      & GAE Lambda ($\lambda_{\text{GAE}}$)                                          & 0.95                            \\
      & Entropy Coefficient for $\pi_{\text{comm}}$                                  & 0.01                            \\
      \midrule
      \multicolumn{3}{l}{\textbf{Policy Network ($\pi_{\text{comm}}$)}}                                                                      \\
      & Transformer Layers ($N_{\pi,\text{enc}}$)                                    & 4                               \\
      & Attention Heads ($H_{\pi,\text{enc}}$)                                        & 4                               \\
      & Model Dimension ($d_{\pi,\text{model}}$)                                     & 256                             \\
      & Feed-Forward Network Dim. ($d_{\pi,\text{ff}}$)                              & 1024                            \\
      \midrule
      \multicolumn{3}{l}{\textbf{CKM ($f_{\text{CKM}}$, $f_{\text{update}}$)}}                                                            \\
      & Transformer Layers in $f_{\text{CKM}}$ ($N_{\text{ckm},\text{enc}}$)         & 2                               \\
      & Attention Heads in $f_{\text{CKM}}$ ($H_{\text{ckm},\text{enc}}$)             & 2                               \\
      & Model Dimension ($d_{\text{ckm},\text{model}}$)                              & 128                             \\
      & GRU Hidden Size in $f_{\text{update}}$ ($d_{\text{gru}}$)                    & 128                             \\
      & History Length for CKM input ($k_{\text{hist}}$)                             & 5 utterances                    \\
      & Learning Rate (Adam, $\alpha_{\text{ckm}}$) for CKM fine-tuning               & $5\times10^{-5}$                \\
      \midrule
      \multicolumn{3}{l}{\textbf{Cognitive Gap Function ($f_{\text{gap}}$)}}                                                               \\
      & Architecture                                                                  & MLP (2 layers, 128 units, ReLU) \\
      & Input Projection Dim.                                                        & 128                             \\
      & Output Gap Vector Dim. ($d_{\text{gap}}$)                                     & 64                              \\
      & Learning Rate (Adam, $\alpha_{\text{gap}}$) for fine-tuning                   & $5\times10^{-5}$                \\
      \midrule
      \multicolumn{3}{l}{\textbf{Reward Function}}                                                                                          \\
      & Communication Cost Weight ($\lambda_{\text{cost}}$)                           & 0.001                           \\
      & Intrinsic Shaped Reward ($r_{\text{shape}}$)                                  & 0.05                            \\
      \midrule
      \multicolumn{3}{l}{\textbf{General Training Setup}}                                                                                   \\
      & Communication Rounds per Episode ($N_{\text{round}}$)                        & 3–5 (curriculum or fixed)       \\
      & Total Training Timesteps                                                     & $5\times10^6$ to $1\times10^7$  \\
      & Base Sentence Encoder                                                        & Sentence-BERT                   \\
      & Linguistic Realization Engine ($f_{\text{LLM}}$)                             & GPT-4 Series / Equivalent API   \\
      \bottomrule
    \end{tabular}%
  }
\end{table}

\begin{table}[htbp]
  \centering
  \caption{Supplementary Hyperparameters for the OSC Framework.}
  \label{tab:hyperparams_osc_supp}
  \resizebox{0.95\columnwidth}{!}{
    \begin{tabular}{@{}llc@{}}
      \toprule
      \textbf{Component Group} & \textbf{Parameter}                & \textbf{Value}                        \\
      \midrule
      \multicolumn{3}{l}{\textbf{State Representation}}                                   \\
      & Embedding Projection Dimension  & 128                                   \\
      & Dialogue History Encoder        & Hierarchical (2 layers, 128 units)    \\
      & History Aggregation Length ($k_h$) & 5 utterances                       \\
      \midrule
      \multicolumn{3}{l}{\textbf{Reward Function}}                                         \\
      & Task Performance Reward ($R_{\text{task}}$) & Success: +1, Failure: –0.1   \\
      & Intrinsic Reward Trigger        & Learned gap resolution               \\
      \midrule
      \multicolumn{3}{l}{\textbf{Policy Network ($\pi_{\text{comm}}$)}}                    \\
      & Discrete Action Space Size      & 10 objectives (extensible)           \\
      & Continuous Style Parameter Range & [0, 1] (uniform)                    \\
      \midrule
      \multicolumn{3}{l}{\textbf{CKM Pre-training}}                                        \\
      & Pre-training Dataset Size       & 1 M dialogue turns                   \\
      & Pre-training LR ($\alpha_{\text{pretrain}}$) & $1\times10^{-4}$            \\
      & Pre-training Objective Weights  & Equal (masked utterance, dialogue act) \\
      \midrule
      \multicolumn{3}{l}{\textbf{Linguistic Realization ($f_{\text{LLM}}$)}}               \\
      & Prompt Length Limit             & 512 tokens                           \\
      & Generation Temperature          & 0.7                                   \\
      & Top-p Sampling                  & 0.9                                   \\
      \bottomrule
    \end{tabular}%
  }
\end{table}

{\footnotesize \textit{Note: The learning rates for CKM ($\alpha_{\text{ckm}}$) and $f_{\text{gap}}$ ($\alpha_{\text{gap}}$) modules during end-to-end fine-tuning are typically set lower than the main policy learning rate $\alpha_{\pi}$ to ensure stability, as these components influence the state representation itself. The specific values are subject to empirical tuning.}}
\section{OSC Hyperparameter Tuning on AlpacaEval 2.0} We tuned the OSC framework on the AlpacaEval 2.0 development set by optimizing communication rounds ($N_{\text{round}}$) and communication cost weight ($\lambda_{\text{cost}}$) to identify the optimal configuration, demonstrating their critical impact on task success rate (LC win rate) and communication efficiency (rounds, token count). Hyperparameter Selection: Communication Rounds ($N_{\text{round}}$): Defines the number of dialogue rounds for agent collaboration, determining interaction depth. Candidate values: \{2, 3, 4, 5\}, covering the default range (3--5). Reason: $N_{\text{round}}$ affects collaboration quality; too few rounds lead to insufficient information, while too many increase redundancy. Communication Cost Weight ($\lambda_{\text{cost}}$): Defines the penalty weight for message token count in the PPO reward function, $\mathcal{R} = R_{\text{task}} - \lambda_{\text{cost}} \cdot C_{\text{comm}}$. Candidate values: \{0.0005, 0.001, 0.002\}, centered on the default 0.001. Reason: $\lambda_{\text{cost}}$ controls communication conciseness, balancing information completeness. Experimental Setup: Dataset: AlpacaEval 2.0 (805 instructions), using development set (~160 instructions) for tuning. Models: Six open-source LLMs (Qwen2-72B-Instruct, LLaMa-3-70B-Instruct, WizardLM-2-8x22B, Gemma2-27B, Deepseek-V3, Deepseek-R1), with Qwen2-72B-Instruct as aggregator. Training: Each configuration trained for $5 \times 10^6$ steps using PPO, with discount factor $\gamma=0.99$ (default). Evaluation Metrics: Task Success Rate: LC win rate (\%), based on GPT-4 evaluator. Communication Efficiency: Average rounds (Avg. Rounds, lower is better), Average token count (Avg. Tokens, k, lower is better). Tuning Method: Grid search (4 $\times$ 3 = 12 configurations), each run 3 times, averaged. Experimental Procedure: Used default configuration ($N_{\text{round}}=4$, $\lambda_{\text{cost}}=0.001$) as baseline. Tested all combinations on the development set, recording LC win rate and communication efficiency. Selected the configuration with the highest LC win rate and reasonable rounds and token count ~\ref{fig:chaocan}.
\begin{figure}[!htbp]
    \centering
    \includegraphics[width=0.95\columnwidth]{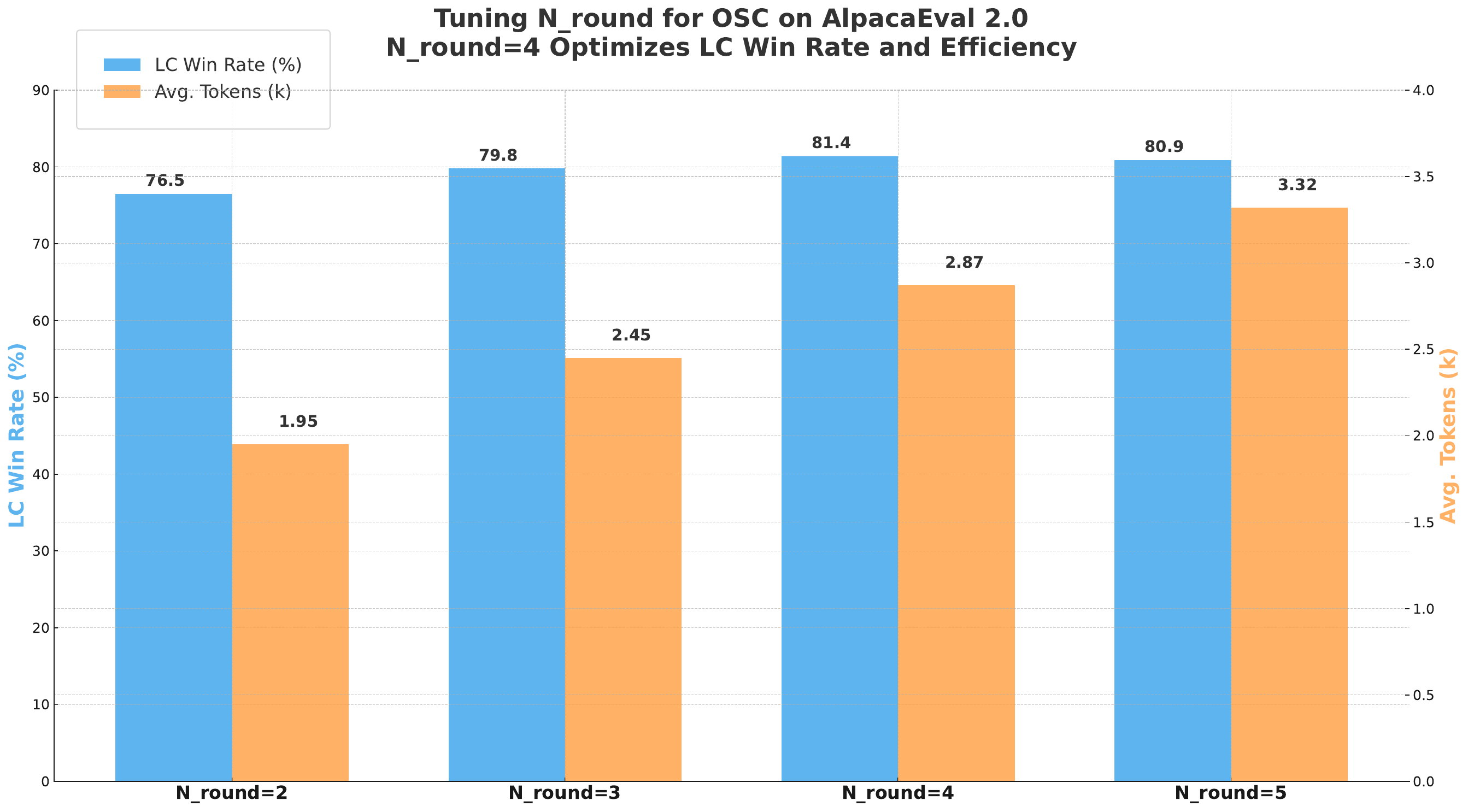}
    \caption{Hyperparameter tuning for communication rounds ($N_{round}$) on AlpacaEval 2.0 shows that $N_{round}=4$ achieves the optimal balance between task success (LC Win Rate) and communication cost (Avg. Tokens).}
    \label{fig:chaocan}
\end{figure}

\section{Reward Function Component Analysis: Detailed Validation of the OSC Framework on AlpacaEval 2.0}
Analyzing the contribution of different components (task reward $R_{\text{task}}$, communication cost $C_{\text{comm}}$, intrinsic shaping reward $r_{\text{shape}}$) in the OSC framework's reward function to collaborative behavior, and detailedly evaluating the impact of each component on task success rate and communication efficiency. The experimental design is as follows: The reward function is formulated as $\mathcal{R} = R_{\text{task}} - \lambda_{\text{cost}} \cdot C_{\text{comm}} + r_{\text{shape}}$. Here, $R_{\text{task}}$ is the task success reward, +1 for success and -0.1 for failure. $C_{\text{comm}}$ is the communication cost (number of message tokens), with $\lambda_{\text{cost}}=0.001$. $r_{\text{shape}}$ is the intrinsic shaping reward (0.05), rewarding the reduction of cognitive discrepancies or the achievement of collaborative goals. Reward combinations include: Only $R_{\text{task}}$, i.e., using only the task reward; $R_{\text{task}} - \lambda_{\text{cost}} \cdot C_{\text{comm}}$, i.e., adding a communication cost penalty; Full Reward ($R_{\text{task}} - \lambda_{\text{cost}} \cdot C_{\text{comm}} + r_{\text{shape}}$), i.e., adding the intrinsic shaping reward. The baseline is KABB, with an LC win rate of 77.9\% (Table 1) and no dynamic communication. Experimental Settings: The dataset used is AlpacaEval 2.0 (containing 805 instructions), with its development set (approx. 160 instructions) used for training and the validation set (approx. 160 instructions) for evaluation. Six open-source LLMs were selected (e.g., Qwen2-72B-Instruct, LLaMa-3-70B-Instruct, etc.), with Qwen2-72B-Instruct serving as the aggregator. Training was conducted using the PPO algorithm for $5 \times 10^6$ environment steps, with $N_{\text{round}}=4$. Evaluation metrics include: Task Success Rate (LC Win Rate, \%); Communication Efficiency, specifically including Average Rounds (Avg. Rounds, lower is better), Average Tokens (Avg. Tokens, in k, lower is better), Communication Redundancy (Redundancy, \%), and Conflict Resolution Rate (Conflict Res., \%). Experimental Procedure: First, initialization is performed by loading the pre-trained CKM and $f_{\text{gap}}$. Then, reward combination experiments are conducted: for each reward combination, OSC is trained on the development set, and CKM, $f_{\text{gap}}$, and $\pi_{\text{comm}}$ are fine-tuned end-to-end. Finally, testing is performed on the validation set, and the metrics are recorded. The experimental results are shown in the table below:
\begin{table*}[htbp]
    \centering
    {\large
    \resizebox{\textwidth}{!}{%
        \begin{tabular}{lccccc}
        \hline
        Reward Combination & LC Win Rate (\%) & Avg. Rounds & Avg. Tokens (k) & Redundancy (\%) & Conflict Res. (\%) \\
        \hline
        Only $R_{\text{task}}$ & 74.1 & 5.9 & 3.95 & 18.9\% & 82.6\% \\
        $R_{\text{task}} - \lambda_{\text{cost}} \cdot C_{\text{comm}}$ & 78.2 & 5.0 & 3.20 & 15.7\% & 86.5\% \\
        Full Reward ($R_{\text{task}} - \lambda_{\text{cost}} \cdot C_{\text{comm}} + r_{\text{shape}}$) & 81.4 & 4.3 & 2.87 & 12.6\% & 91.7\% \\
        KABB (Baseline) & 77.9\% & - & - & - & - \\
        \hline
        \end{tabular}%
    }}
    \caption{Analysis of the reward function components, showing that the full reward (including task success, communication cost, and intrinsic shaping) achieves the best performance and communication efficiency compared to simpler reward structures and the KABB baseline.}
    \label{tab:resizebox}
\end{table*}
Results Analysis: When using only $R_{\text{task}}$, the LC win rate was 74.1\%, lower than KABB's 77.9\%, mainly due to a lack of guidance for collaboration. At this point, Avg. Rounds was 5.9, Avg. Tokens was 3.95k, Redundancy was 18.9\%, and Conflict Res. was 82.6\%, indicating low communication efficiency. After introducing $R_{\text{task}} - \lambda_{\text{cost}} \cdot C_{\text{comm}}$, the LC win rate increased to 78.2\%, close to KABB. The communication cost penalty effectively reduced the number of rounds (5.0) and tokens (3.20k). Redundancy decreased to 15.7\%, and Conflict Res. improved to 86.5\%, indicating some improvement in collaborative behavior. With the full reward, the LC win rate reached 81.4\% (Table 1), outperforming KABB. Avg. Rounds decreased to 4.3, Avg. Tokens to 2.87k, Redundancy to 12.6\%, and Conflict Res. increased to 91.7\%, demonstrating optimal collaborative performance. The KABB baseline had an LC win rate of 77.9\% but no relevant data on dynamic communication. Further Analysis: When using only $R_{\text{task}}$, the sparse reward led to slow learning of collaborative behavior, resulting in a lower LC win rate (74.1\%) and more redundant communication (18.9\%). After adding $C_{\text{comm}}$, the communication cost penalty encouraged the model to generate more concise communication, reducing rounds from 5.9 to 5.0, tokens from 3.95k to 3.20k, and increasing the LC win rate from 74.1\% to 78.2\%. After adding $r_{\text{shape}}$, the intrinsic shaping reward effectively guided collaborative behavior (e.g., promoting the reduction of cognitive discrepancies), leading to an LC win rate of 81.4\%, an increase in conflict resolution rate to 91.7\%, and a decrease in communication redundancy to 12.6\%. Compared to KABB, the OSC framework with the full reward outperformed KABB in LC win rate (81.4\% vs. 77.9\%), indicating that the dynamic reward mechanism achieved significant effects.
\section{OSC Computational Resource Efficiency Results}
We adopt the AlpacaEval 2.0 dataset (160 development examples, 160 validation examples), six agents (e.g., LLaMa-3-13B-Instruct and other compressed models) with a Qwen2-13B aggregator in the OSC system, running on a single NVIDIA A100 GPU. Training uses mixed precision for $1\times10^{6}$ steps, freezing the CKM and $f_{\mathrm{gap}}$ modules and training only $\pi_{\mathrm{comm}}$. During inference, we apply INT8 quantization, set $N_{\mathrm{round}} = 3$, and cache CKM states. Hyperparameters are $N_{\mathrm{round}} = 3$, $\lambda_{\mathrm{cost}} = 0.001$, and $\gamma = 0.99$. We evaluate training GPU hours, training memory usage (GB), inference latency (seconds per instruction), inference memory usage (GB), and LC win rate (\%). As shown in Table~\ref{tab:osc_model_comparison} , OSC requires 10.8 GPU hours for training, uses 11.3 GB of memory during training, achieves 1.79 s per instruction and 7.8 GB of memory during inference, and attains an LC win rate of 78.6\%.

\end{document}